\documentclass{egpubl}

\JournalPaper         
\usepackage[T1]{fontenc}
\usepackage{dfadobe}  

\usepackage{cite}  
\BibtexOrBiblatex
\electronicVersion
\PrintedOrElectronic
\ifpdf \usepackage[pdftex]{graphicx} \pdfcompresslevel=9
\else \usepackage[dvips]{graphicx} \fi

\usepackage{egweblnk} 

\usepackage{amsmath,amssymb} 
\usepackage{color}
\usepackage{bm}
\usepackage{makecell}
\usepackage{float}
\usepackage{diagbox}
\usepackage{CJKutf8}
\usepackage[colorlinks,linkcolor=blue]{hyperref}

\title[Self-Supervised Learning of Part Mobility from Point Cloud Sequence]%
{Self-Supervised Learning of Part Mobility \\ from Point Cloud Sequence}

\author[Y. Shi et al.]
{\parbox{\textwidth}{\centering Yahao Shi, Xinyu Cao and Bin Zhou\thanks{Corresponding author: zhoubin@buaa.edu.cn}
}
\\
{\parbox{\textwidth}{\centering State Key Laboratory of Virtual Reality Technology and Systems, School of Computer Science and Engineering, Beihang University, Beijing, China
}
}
}

\begin{document}


\maketitle
\begin{abstract}
Part mobility analysis is a significant aspect required to achieve a functional understanding of 3D objects. It would be natural to obtain part mobility from the continuous part motion of 3D objects. In this study, we introduce a self-supervised method for segmenting motion parts and predicting their motion attributes from a point cloud sequence representing a dynamic object. To sufficiently utilize spatiotemporal information from the point cloud sequence, we generate trajectories by using correlations among successive frames of the sequence instead of directly processing the point clouds. We propose a novel neural network architecture called PointRNN to learn feature representations of trajectories along with their part rigid motions. We evaluate our method on various tasks including motion part segmentation, motion axis prediction, and motion range estimation. The results demonstrate that our method outperforms previous techniques on both synthetic and real datasets. Moreover, our method has the ability to generalize to new and unseen objects. It is important to emphasize that it is not required to know any prior shape structure, prior shape category information, or shape orientation. To the best of our knowledge, this is the first study on deep learning to extract part mobility from point cloud sequence of a dynamic object.


\end{abstract}  
\section{Introduction}

In the real world, there often exist a number of dynamic and articulated objects, which can be directly operated using their moving parts. For example, we may need to open a refrigerator door to keep or remove an object. If we expect autonomous agents to interact with such objects correctly, they must have the ability to analyze which part of an object is a moving part. Hence, learning part mobility of 3D objects is beneficial for 3D computer vision~\cite{Li_CGF2016} and robotics~\cite{Hermans_2013} and is closely related to the understanding of object affordances~\cite{Myers_ICRA2015}, functionality~\cite{Hu_CGF2018}, and interaction using object recognition~\cite{Liu_2019_CVM} and human motion capture data~\cite{Roberts_2019_CVM}.

Recently, with the emergence of large 3D labeled datasets and deep learning techniques, several studies have been conducted to make considerable progress in supervised semantic part segmentation, such as PointNet++~\cite{Pointnet++} and PointCNN~\cite{PointCNN}. However, these segments were often defined by personal subjective experience, which can easily lead to ambiguity. Thus, research on learning moveable part segmentation of 3D objects is of more practical significance, and it helps agents understand the essential features of dynamic objects. Moreover, current research on parsing 3D shape representations into semantic parts faces challenges in processing novel object categories and discovering new functional parts. It is not conducive for agents to explore strange new worlds. Ideally, intelligent agents should be able to parse 3D shapes into previously unseen functional parts from observations of continuous part motion. Mobility-based shape parsing brings a novel perspective to the part determination problem and provides an unambiguous decomposition. Additionally, a mobility-based part segmentation algorithm enables intelligent agents to make better use of man-made objects designed to function or interact with other objects (including humans).

\begin{figure}[t]
\centering
\includegraphics[width= \linewidth]{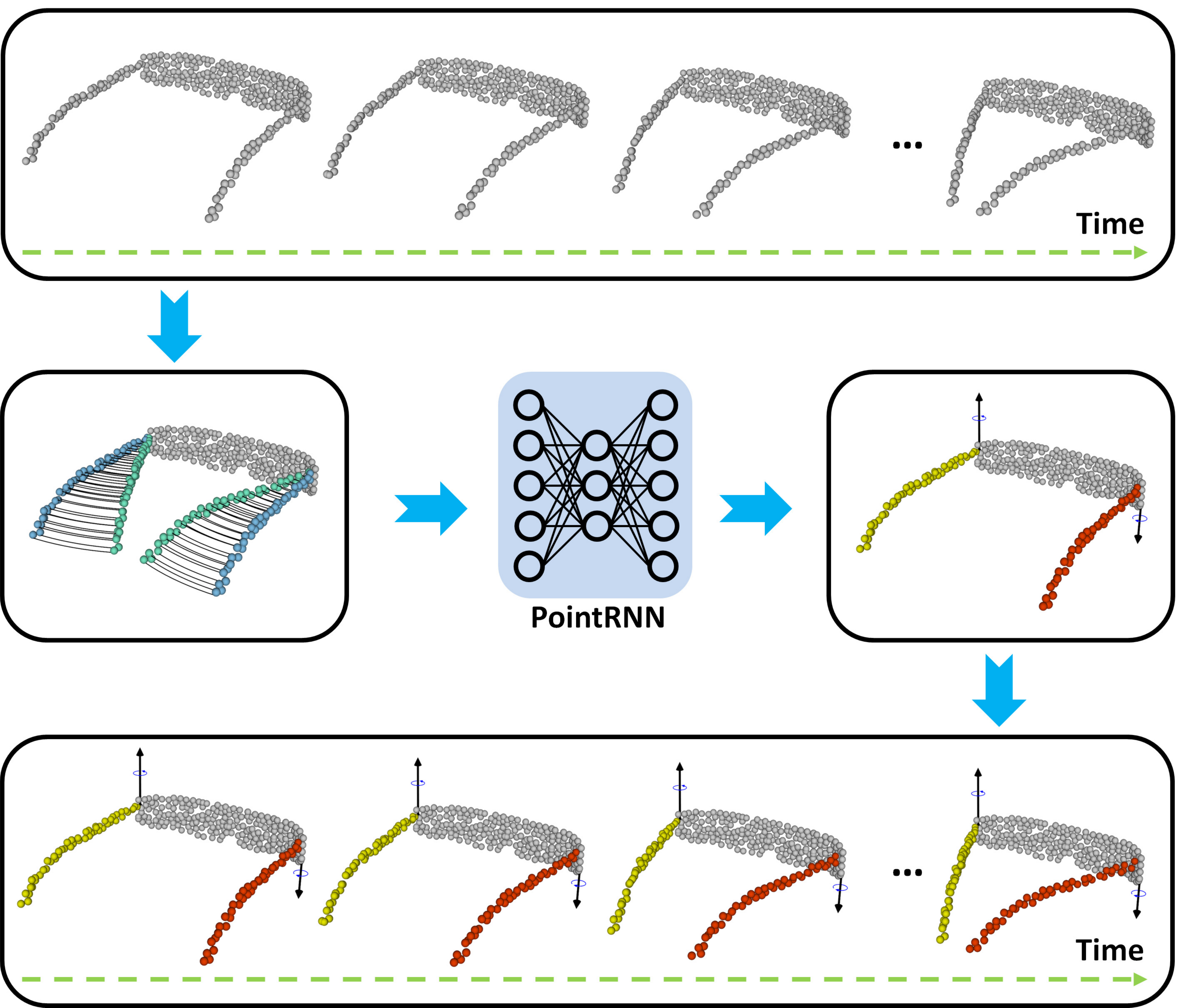}
\caption{In our method, we first convert a raw point cloud sequence into trajectories. Then, we adopt PointRNN to extract varieties of part rigid motion hypotheses by taking these trajectories as input. Finally, we merge these hypotheses and segment motion parts using an iterative algorithm. The results can propagate from the first frame to other frames.}
\label{fig:teaser}
\end{figure}

In this study, we are interested in discovering the part mobility of 3D objects by observing their continuous part motion. We define part mobility as motion part segmentation, motion axis prediction, and motion range estimation. In previous studies, motion parts have been typically extracted from one or two single static snapshots of dynamic objects. By contrast, we aim to deduce motion part structure from observations of continuous articulation states of an object. The reason is that speculating the motion parts and confirming motion vectors by observing only a few motion states can easily cause confusion. Moreover, point cloud sequences reveal sufficient spatiotemporal information, and they are easier to obtain than large well-annotated datasets, owing to recent advances in the techniques of real-time 3D acquisition such as commercial RGB-D cameras. Therefore, we adopt point cloud sequences rather than a single point cloud to represent dynamic objects.

Automatic motion part induction from point cloud sequences is challenging for several reasons. First, objects differ significantly because of their geometry and pose. Motion directions and motion ranges of motion parts have noticeable differences. Second, a point cloud is unordered. There is no tight point correspondence between adjacent frames of point cloud sequences. Third, we must consider the influence factors of the acquisition quality of scan data, including noisy and missing data.

To address the above mentioned problems, we propose a novel self-supervised deep neural network-based method (See Fig.~\ref{fig:teaser}), inspired by the traditional methods of Yan and Pollefeys~\cite{Yan_ECCV2006}. We observed that points on the same part produce similar motion trajectories. Therefore, we treat motion part segmentation as a trajectory clustering problem, and we can calculate rigid motion hypotheses based on these trajectory sets. However, directly handling point cloud sequences can be difficult in a network without point correspondence. Thus, we first transform a point cloud sequence into a bunch of trajectories. Then, we design a neural network called PointRNN to process these trajectories, and it can learn latent trajectory feature representations and generate candidate rigid motion hypotheses represented by motion axes and motion ranges. Finally, we merge the trajectories belonging to a similar rigid motion to achieve motion part segmentation.

Our method is verified through qualitative and quantitative evaluations on both synthetic and real datasets. Additionally, we conduct ablation experiments to confirm the influence of different loss designs, hyperparameter settings, etc. The experimental results demonstrate that our method achieves better performance and requires less time than the traditional method~\cite{Yuan_CGF2016} and deep learning methods~\cite{Wang_CVPR2019,Yi_SIGa2018,Yan_SIGa19}. Moreover, results also demonstrate that our approach can tolerate some noises and has the ability to apply to unseen categories.

In summary, a new self-supervised approach is proposed to parse 3D shapes into moving parts, motion axes, and motion ranges from point cloud sequences without labeling or any prior knowledge. Specifically, our method makes three key contributions:
\begin{itemize}
\item We introduce a self-supervised approach to learning motion part segmentation, motion attribute estimation, and motion flow prediction.
\item We propose a novel neural network called PointRNN, which has the capability to process trajectories and extract a feature representation.
\item Experimental results demonstrate that the performance of our network is superior to state-of-the-art methods, and it can apply to unseen object categories.
\end{itemize}

\section{Related Work}
Deformable shape registration~\cite{ARAP} is a fundamental problem in computational geometry and widely applied to various fields such as computer vision. There are various deformable registration algorithms for point clouds. Many methods are extensions of the classic ICP algorithm~\cite{BM92}. Papazov and Burschka~\cite{Papazov_and_Burschka_CGF2011} proposed an algorithm that computes shape transitions based on local similarity transforms, thereby allowing it to model not only as-rigid-as-possible deformations but also shapes on local and global scales.

Many 3D shape segmentation approaches have been proposed in previous studies to extract moving parts from an RGB-D sequence, a single point cloud, a point cloud sequence, and a mesh model. Given a specified RGB-D sequence that contains a dynamic object, attempts have been made in previous studies to recover the 3D scene flow, thus discovering moving parts of 3D objects. Jaimez et al.~\cite{Jaimez_ICRA2015} proposed a primal--dual algorithm to compute RGB-D flow for estimating heterogeneous and non-rigid motion at a high frame rate. Vogel et al.~\cite{Vogel_ECCV2014} introduced a method to recover dense 3D scene flow from multiple consecutive frames in a sliding temporal window. Motion part segmentation can be extracted by point correspondence, as shown by Fayad et al.~\cite{Fayad_ICCV2011}, and then they can utilize these motion parts to reconstruct the articulated structure. However, common defects in these approaches are their reliance on RGB color to compute scene flow and inability to handle complex structures or large motions.

In the case of a raw 3D point cloud sequence, many studies aimed to establish a point-wise correspondence between consecutive point clouds of an articulated shape~\cite{Chang_and_Zwicker_CGF2008},~\cite{Papazov_and_Burschka_CGF2011}. Yan and Pollefeys~\cite{Yan_ECCV2006} cast the problem of motion segmentation of feature trajectories as linear manifold finding problems and proposed a general framework for motion segmentation under affine projections. Besides, Kim et al.~\cite{Kim_IROS2016} considered that trajectories could be grouped by clustering to separate different motion parts. Yuan et al.~\cite{Yuan_CGF2016} proposed a local-to-global approach to co-segment point cloud sequences of articulated objects into near-rigid moving parts. Most of the methods mentioned above require a large amount of computation to achieve better performance. These approaches also require considerable effort in threshold setting from case to case.

Several approaches parse mesh models into moving parts. Mitra et al.~\cite{Mitra_2013} inferred the motion of individual parts and the interactions among parts based on their geometry and a few user-specified constraints. They utilized the results to illustrate the motion of mechanical assemblies. Hu et al.~\cite{Hu_SIGa2017} introduced a data-driven approach for learning part mobility based on a defined motion pattern from a single static state of 3D objects. However, these methods are based on a well-defined motion pattern and well-segmented 3D objects.

Recent supervised learning approaches in 3D shape segmentation employ deep network architecture to train a classifier on labeled data represented by a point cloud~\cite{Pointnet++}, volumetric grids~\cite{Maturana_and_Scherer_ICRA2015} or spatial data structures~\cite{Klokov_ICCV2017}. Subsequent attempts have been made to extract part rigid motions using new deep learning technology. Wang et al.~\cite{Wang_CVPR2019} proposed the shape2motion network architecture, which takes a single point cloud as input. Their network aims to simultaneously segment motion parts and predict motion axes based on a large, well-annotated dataset. However, their approach has difficulty in discovering new motion parts and generalizing them to novel categories. Their network requires a considerable amount of time for supervised learning using a large dataset. Another type of method, such as that proposed by Yi et al.~\cite{Yi_SIGa2018}, infers part motion flow and estimates part correspondence by comparing two different motion states. Unfortunately, they did not perform a specific analysis of motion attributes such as motion axes and motion ranges. Behl et al.~\cite{Behl_CVPR2019} and Liu et al.~\cite{Liu_CVPR2019} promoted this class of algorithm to identify motion flow in scene data. However, these methods still rely on supervised learning. Yan et al.~\cite{Yan_SIGa19} introduced RPM-Net to infer movable parts of a single point cloud. They adopted a recurrent neural network (RNN) to movable segment parts by forecasting a temporal sequence motion of 3D objects. In contrast, our network takes point cloud sequences as input instead of a single point cloud. Furthermore, we employ an RNN to process each trajectory locally rather than the entire point cloud directly and globally.

\section{Method}
Our method comprises three steps for learning part mobility, as shown in Fig.~\ref{fig:teaser}. First, we adopt a deformable registration algorithm using point correspondence to generate trajectories (Sec.~\ref{sec:section3_1}). Second, we propose a network architecture called PointRNN to produce candidate motion hypotheses extracted from trajectories (Sec.~\ref{sec:section3_2}). Third, we design an iterative algorithm for removing redundant motion hypotheses and segmenting a point cloud into several motion parts while considering the matching degree between trajectories and motion hypotheses (Sec.~\ref{sec:section3_3}). The second step of our method occurs within the learning pipeline, and the other two steps occur outside it. In this section, we discuss the modules of our approach in detail.

\subsection{Trajectory Generation}
\label{sec:section3_1}
Two adjacent frames of the raw point cloud sequence do not have point-wise correlations because of unordered points. Therefore, we can not directly concatenate all the raw point cloud sequence frames and take them as input to the neural network. To employ the deep learning approach, we first convert the raw point cloud sequence into trajectories. Moreover, we adopt the method of Papazov and Burschka~\cite{Papazov_and_Burschka_CGF2011} based on local similarity transforms to generate motion trajectories to ensure a fair comparison with Yuan et al.~\cite{Yuan_CGF2016}, and this method is an alternative module to generate trajectories for our pipeline. All the trajectories are cropped to an equal length to facilitate deep neural network processing.

Given two adjacent frames, $PC_{s}$ and $PC_{t}$, of a point cloud sequence, the registration algorithm includes iterative steps of correspondence and deformation. In terms of the correspondence step, we find the closest point in $p' \in PC_{t}$ for each point $p \in PC_{s}$. We optimize $p'$ by searching over its $k$-nearest neighbors to make the match error smooth. Formally, we compute a mapping, $\pi_{s \rightarrow t}$, by minimizing the Laplacian smoothness energy of the residual field $\{ \pi_{s \rightarrow t}(p)-p'\}$:

\begin{equation}
\mathcal{E}_{smooth} (\pi_{s \rightarrow t}) = \sum_{p \in PC_{s}} \lVert \delta(p) \rVert^{2},
\end{equation}

where the residual Laplacian, $\delta(p)$, is defined as

\begin{equation}
\delta(p) = \frac{1}{\left|\mathcal{N}(p)\right|}\sum_{q\in \mathcal{N}(p)} [(\pi_{s \rightarrow t}(p)-p')-(\pi_{s \rightarrow t}(q)-q')].
\end{equation}

Here, $\mathcal{N}(p)$ is the $k$-nearest neighbors of $p$. In terms of the deformation step, we minimize a fitting error between the local neighborhood of $p\in PC_{s}$ and its matched counterpart to estimate a similarity transformation, denoted as $(\bm{R},\bm{t})$, by using the singular value decomposition~\cite{Scott_SIGGRAPH2006}.

\begin{equation}
\mathcal{E}_{fitting}(\bm{R},\bm{t}) = \sum_{q\in \mathcal{N}(p)} \lVert \pi_{s \rightarrow t}(q) - (\bm{R}q + \bm{t}) \rVert^{2}.
\end{equation}

\begin{figure*}[t]
\begin{center}
\includegraphics[width=1.0\linewidth]{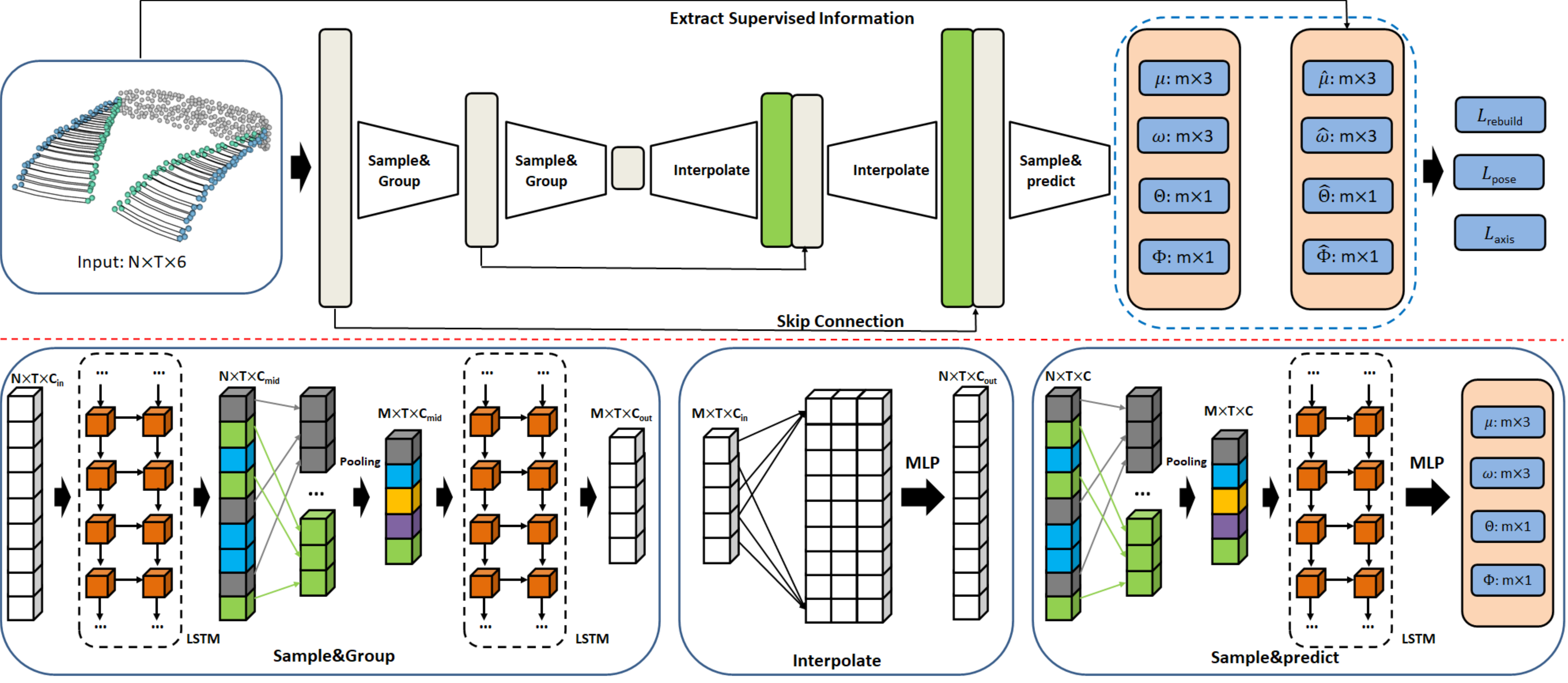}
\end{center}
\caption{PointRNN network architecture consists of three parts. The Sample\&Group Module is employed to learn temporal and spatial features from trajectories. We adopt the Interpolate Module to recover original resolution features from global features. We feed these features to the Sample\&Predict Module to estimate motion parameters. Our supervised information is automatically obtained from the input. $c_{in}$, $c_{mid}$ and $c_{out}$ are the number of channels, which are different in different layers. Specifically, $c_{in}=6,c_{mid}=128,c_{out}=64$ in the first Sample\&Group Module. $c_{in}=64,c_{mid}=128,c_{out}=256$ in the second Sample\&Group Module. These are hyperparameters in our method.}
\label{fig:network}
\end{figure*}

\subsection{Part Rigid Motion Hypothsis Extraction}
We propose a novel neural network called PointRNN, which has the capacity to process trajectories and extract the feature representation, as illustrated in Fig.~\ref{fig:network} (Sec. \ref{sec:section_3_21}). Moreover, we design a self-supervised loss function to train the PointRNN (Sec.~\ref{sec:section_3_22}).
\label{sec:section3_2}
\subsubsection{PointRNN Network Architecture}
\label{sec:section_3_21}
Given a set of trajectories, $\{\tau_{1}, \tau_{2}, ..., \tau_{n}\}$ ($n=2048$ by default), $\tau_{i}$ has a point set,  $\{x_{i,1}, x_{i,2}, ..., x_{i,t}\}$ ($t=11$ by default), with $x_{i,j} \in \mathbb{R}^3$ and a motion direction set, $\{d_{i,1}, d_{i,2}, ..., d_{i,t}\}$ ($t=11$ by default), with $d_{i,j} \in \mathbb{R}^3$. There is a one-to-one relationship between these two sets, such as $(x_{i,1}, d_{i,1})$. For each trajectory, we employ a Long Short-Term Memory (LSTM) network~\cite{LSTM} to encode time information, because LSTM has demonstrated high accuracy in tasks related to the processing of sequential and time-series data. In previous work, Yan et al.~\cite{Yan_SIGa19} used the PointNet++ encoder to create a global feature for the input point cloud. Then, they fed this feature vector to the LSTM network. Compared to their approach, ours pays more attention to local and detailed features for each trajectory. 

We assume that similar trajectories should belong to one motion part, as inspired by the traditional clustering method~\cite{Yan_ECCV2006}. Thus, we design a Sample\&Group Module to learn the spatial and temporal features with their underlying motion from these similar trajectories. To this end, our network need to partition the set of trajectories into overlapping local similar trajectory partitions by the distance metric and then learns their features. Therefore, our network first performs a downsampling process on the set of trajectories. We adopt iterative farthest point sampling to obtain a subset of trajectories, $\{\tau_{1}, \tau_{2}, ..., \tau_{i}\}$, such that $\tau_{j}$ is the most distant trajectory (in metric distance) from set $\{\tau_{1}, \tau_{2}, ..., \tau_{j-1}\}$ ($j \leq i$). The advantage of the farthest point sampled algorithm is that it can cover the trajectory set in the space. Moreover, the sampling trajectories are isometry-invariant. The farthest point sampling algorithm is widely used in deep learning to process point cloud, such as PointNet++~\cite{Pointnet++}. The metric distance function between two trajectories, $(\tau_{i}, \tau_{j})$, is as follows:

\begin{equation}
dis_{(\tau_i,\tau_j)} = \frac{1}{t}\sum_{1 \leq m \leq t} \lVert x_{i,m} - x_{j,m} \rVert^2.
\label{trajectory_dis}
\end{equation}

We adopt Euclidean distance rather than Hausdorff distance or Fréchet distance because the latter two metrics require significantly more computations without yielding improved performance. Next, we search a neighborhood trajectory set, $\{\tau_{i_1}, \tau_{i_2}, ..., \tau_{i_k}\}$, for each sampled trajectory, $\tau_i$, using the $k$-nearest neighbor method, where the metric distance between $\tau_i$ and $\tau_{i_m}$ ($1 \leq m \leq k$) does not exceed a certain threshold $\epsilon$. Then, we obtain several overlapping partitions of trajectories. 

Next, our network learns motion features from these partitions. Specifically, we employ an LSTM network to encode time information for each point trajectory. This feature contains the positions of each point at different times and the movement trend of each point. We utilize two LSTM networks before and after sampling. To encode spatial information, we aggregate the features of the trajectory partition by a max-pooling operation. This feature contains all trajectory features from each trajectory partition. We treat this feature as regional spatial features and express them as features of $\tau_i$. In our method, $k$ and $\epsilon$ are hyperparameters in different layers.

We can obtain a global feature after Sample\&Group Module, and then we need to propagate global features to single trajectory features. To this end, we design an Interpolate Module with distance-based interpolation and skip-connection. We achieve feature propagation by interpolating features from the $k$-nearest trajectory set, $\{\tau_{j_1}, \tau_{j_2}, ..., \tau_{j_k}\}$ ($k = 3$ by default), to each trajectory, $\tau_j$. We adopt the inverse distance-weighted average based on the $k$-nearest neighbors in interpolation. Finally, we feed these interpolated trajectory features into a Multilayer Perceptron (MLP) Network to encode single trajectory features.

\begin{figure}[t]
\centering
\includegraphics[width=\linewidth]{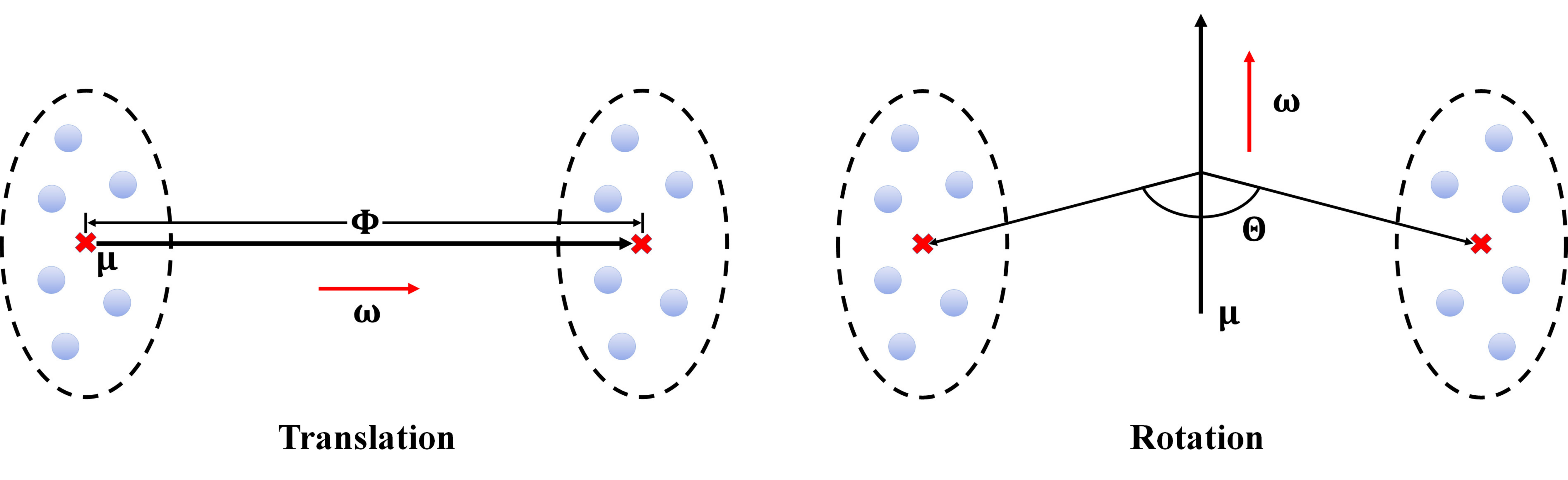}
\caption{Motion parameters ($\mu$, $\omega$, $\Theta$, and $\Phi$). $\mu$ is the start of the motion axis. The motion direction ($\omega$) is a unit vector. $\Phi$ is the translation distance, and $\Theta$ is the rotation angle.}
\label{fig:motion_parameters}
\end{figure}

We can obtain the final trajectory features after Interpolate Module. Then, we need to utilize these features to estimate the motion attribute for each motion part. Our network will produce abundant motion hypotheses because the number of motion part is uncertain for each object. Therefore, we design a Sample\&Predict Module to generate part rigid motion hypotheses. By observing most objects in the real world, we found that the number of motion parts is finite and small. We assume that the number of motion parts is no more than ten based on our dataset. Similar to the Sample\&Group Module, we obtain $m$ ($m = 64$ by default) trajectory partitions using farthest point sampling and $k$-nearest neighbor method ($k = 32$ by default). We consider that these sets can cover all the motion parts of 3D shapes. We utilize a max-pooling operation to obtain aggregated spatial features for each trajectory set. Then, we obtain temporal features using an LSTM network. Finally, we feed their temporal and spatial features to an MLP Network. The network outputs a motion axis including a start point ($\mu$) and direction ($\omega$), motion ranges including a shifted distance ($\Phi$), and a rotation angle ($\Theta$) (Fig. ~\ref{fig:motion_parameters}).

\subsubsection{Self-Supervised Loss Function Design.}
\label{sec:section_3_22}
We assume that trajectories contain rich motion information. Therefore, we can utilize estimated motion parameters to confirm whether it can reconstruct the motion sequence. Moreover, we can calculate the approximate parameters from motion sequence and take them as supervised information.

To reconstruct the motion sequences, we compute the motion matrix, $\bm{M}\in \mathbb{R}^{4\times4}$, by taking advantage of network output. The motion matrix is composed of a rotation matrix and a translation matrix ($\bm{M}=\bm{R}+\bm{t}$). $\bm{t}$ is easy to generate by using $\omega$ and $\Phi$. The problem of solving $\bm{R}$ can be reduced to transforming the point set rotating around the given axis (including $\mu$ and $\omega$) into the point set rotating around the z-axis of the standard coordinate system, $O_{xyz}$. The basic idea of deriving this matrix is to divide the problem into a few known simple steps:
\begin{itemize}
\item We align the start of the given axis with the coordinate origin and move the point set.
\item We rotate the given axis, and the point set such that the axis lies in the $yOz$ coordinate planes.
\item We rotate the given axis, and the point set such that the axis is aligned with the z-axis.
\item We utilize one of the fundamental rotation matrices to rotate the point sets using $\Theta$. Finally, we undo steps three to one.
\end{itemize}

We can rebuild the last frame ($PC_{t}$) from the first frame ($PC_{1}$) by pre-multiplication ($M$ on the left) in our neural network. Moreover, we implement the partial derivative for each network output in our neural network to apply the backpropagation algorithm. Therefore, we define the first loss function, called the rebuild loss function, as follows:

\begin{equation}
\mathcal{L}_{rebuild} = \lVert PC_{t} - PC_{1}\cdot M \rVert^2.
\label{loss_rebuild}
\end{equation}

For each trajectory set in the Sample\&Predict Module, we solve an absolute orientation problem to obtain an approximate solution of rigid motion ($\hat{\bm{M}}$). Our solution for the absolute orientation problem is based on the method of Myronenko and Song~\cite{Myronenko_and_Song_arXiv2009}. Besides, we guarantee that the determinant of $\hat{\bm{M}}$ is 1 to ensure that $\hat{\bm{M}}$ has a rotation matrix other than the reflection matrix. Then, we extract rotation matrix $\bm{R}\in \mathbb{R}^{3\times3}$ from $\bm{M}$. Similarly, we can obtain $\hat{\bm{R}}$ from $\hat{\bm{M}}$. We define the relative pose estimation loss function, which was used to measure the angular distance between $\hat{\bm{R}}$ and $\bm{R}$ in the study by Suwajanakorn et al.~\cite{Supasorn_NIPS2018}.

\begin{equation}
\mathcal{L}_{pose} = 2\arcsin(\frac{1}{2\sqrt{2}}\lVert \hat{\bm{R}} - \bm{R} \rVert^2).
\label{loss_pose}
\end{equation}

We found that the network has a slow convergence and easily falls into a local optimum solution using only $\mathcal{L}_{rebuild}$ and $\mathcal{L}_{pose}$. The reason for this is that it is possible to generate the same $\bm{M}$ by using two different parameter sets of $\mu$, $\omega$, $\Phi$, and $\Theta$. To solve this problem, we compute an approximate solution of $\mu$, $\omega$, $\Phi$, and $\Theta$ denoted as $\hat{\mu}$, $\hat{\omega}$, $\hat{\Phi}$, and $\hat{\Theta}$, respectively.
In general, $\hat{\bm{R}}$ can be written more concisely as Rodrigues’ rotation formula~\cite{RodriguesRotationFormula} as follows:

\begin{equation}
\hat{\bm{R}} = (\cos\hat{\Theta})\textit{I} + (\sin\hat{\Theta})[\hat{\omega}]_\times + (1-\cos\hat{\Theta})(\hat{\omega}\otimes\hat{\omega}).
\label{Rodrigues' rotation formula}
\end{equation} where $[\hat{\omega}]_\times$ is the cross product matrix of $\hat{\omega}$, $\hat{\omega}\otimes\hat{\omega}$ is the outer product, and $\textit{I}$ is the identity matrix.
We can easily obtain $\hat{\Theta}$ and $\hat{\omega}$ from equation (\ref{Rodrigues' rotation formula}).
\begin{equation}
\hat{\Theta} = \arccos(\frac{tr(\hat{\bm{R}})-1}{2}),
\end{equation} where $tr(\hat{\bm{R}})$ is the trace of $\hat{\bm{R}}$.
\begin{equation}
\hat{\omega} = \frac{1}{2\sin\Theta}\begin{pmatrix}
\hat{\bm{R}}[2,1] - \hat{\bm{R}}[1,2]\\
\hat{\bm{R}}[0,2] - \hat{\bm{R}}[2,0]\\
\hat{\bm{R}}[1,0] - \hat{\bm{R}}[0,1]
\end{pmatrix}.
\end{equation}

We extract $\hat{\bm{t}}\in\mathbb{R}^3$ from $\hat{\bm{M}}$, where $\hat{\bm{t}}$ is a vector from the centroid of $PC_1$ to that of $PC_t$. $\hat{\Phi}$ is the length of $\hat{\bm{t}}$ projected to $\hat{\omega}$.
\begin{equation}
\hat{\Phi} = \bigg| \frac{\hat{\omega}\cdot\hat{\bm{t}}}{\lVert \hat{\omega} \rVert^2} \bigg|.
\end{equation}

Finally, we solve linear algebraic equations to obtain $\hat{\mu}$, where $\textit{I}$ is the identity Matrix.
\begin{equation}
\hat{\mu} = (\hat{\bm{R}} - \textit{I})^{-1}\cdot(\hat{\Phi}\cdot\hat{\omega}-\hat{\bm{t}}).
\end{equation}

In the case of the general situation in which the start of an axis is not at the coordinate origin, we can obtain the same result. Then, we define the axis loss function as follows:
\begin{equation}
\begin{split}
\mathcal{L}_{axis} = \lVert \hat{\Theta} - \Theta \rVert^2 + \lVert \hat{\Phi} - \Phi \rVert^2 + cosine(\hat{\omega},\omega) \\
+ \frac{1}{3}\lVert (\hat{\bm{R}}-\textit{I})\cdot\mu - (\hat{\Phi}\cdot\hat{\omega}-\hat{t}) \rVert^2,
\label{loss_axis}
\end{split}
\end{equation} where $cosine(\hat{\omega},\omega)$ is the cosine distance between $\hat{\omega}$ and $\omega$. Given that $(\hat{\bm{R}}-\textit{I})^{-1}$ is difficult to compute, network output $\mu$ should satisfy $(\hat{\bm{R}}-\textit{I})\cdot\mu - (\hat{\Phi}\cdot\hat{\omega}-\hat{t}) = 0$ rather than directly regressing $\mu$ by $\hat{\mu}$.

The final loss function is the weighted sum of $\mathcal{L}_{rebuild}$ (\ref{loss_rebuild}), $\mathcal{L}_{pose}$ (\ref{loss_pose}), and $\mathcal{L}_{axis}$ (\ref{loss_axis}).
\begin{equation}
\mathcal{L}_{total} = \alpha_1\cdot\mathcal{L}_{rebuild} + \alpha_2\cdot\mathcal{L}_{pose} + \alpha_3\cdot\mathcal{L}_{axis}.
\end{equation}

In our implementation, $\alpha_1 = \alpha_2 = \alpha_3 = 1$ by default. In addition, we set a threshold for each loss function to tolerate error and improve robustness, because approximate solutions of $\hat{\mu}$, $\hat{\omega}$, $\hat{\Phi}$, and $\hat{\Theta}$ usually have errors with respect to the ground truth, especially in real scan data.

\begin{figure}[t]
\centering
\includegraphics[width=\linewidth]{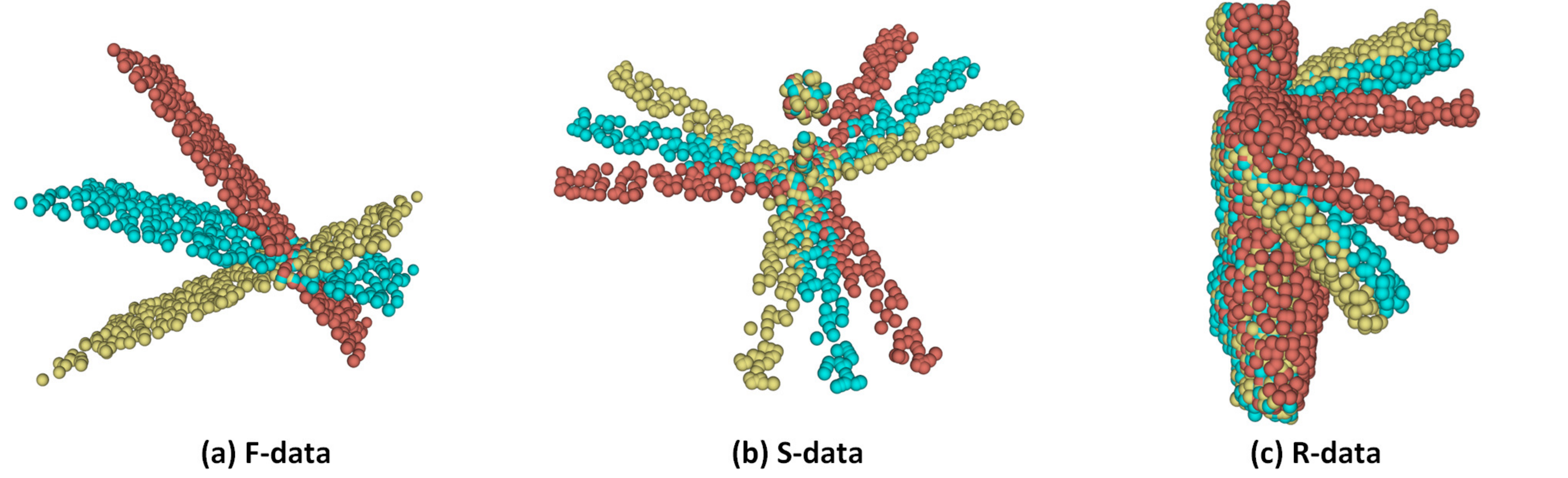}
\caption{Three frames of point cloud sequences in each of the three datasets used in our experiments.}
\label{fig:dataset}
\end{figure}

\subsection{Motion Part Segmentation and Optimization}
\label{sec:section3_3}
However, these hypotheses are redundant, such that we must merge similar motion hypotheses. We define that there is no movement if $\Theta$ and $\Phi$ are no more than $\mathcal{E}_{\Theta}$ (0.1 radians by default) and $\mathcal{E}_{\Phi}$ (0.05 length by default), respectively, considering noisy data and computation errors. Further, we define rotation as $\Theta > \mathcal{E}_{\Theta}$ and $\Phi < \mathcal{E}_{\Phi}$ and translation by contrast. The combination of rotation and translation satisfies the conditions that $\Theta > \mathcal{E}_{\Theta}$ and $\Phi > \mathcal{E}_{\Phi}$. We merge the remaining motion axes by comparing the diversity of $\mu$, $\omega$, $\Phi$, and $\Theta$.

We first define the metric distance between trajectory $\tau_i$ and motion hypotheses as the matching degree to merge motion hypotheses. Therefore, we utilize the motion hypotheses to rebuild the trajectory, denoted as $\hat{\tau}_i$. Then, the metric distance has two terms: rebuild loss and direction cosine distance.
\begin{equation}
dis_{match} = \frac{1}{t}\sum_{1 \leq m \leq t}( \lVert x_{i,m} - \hat{x}_{i,m} \rVert^2 + \alpha\cdot cosine(d_{i,m},\hat{d}_{i,m})),
\end{equation}

where $\alpha$ is 0.2 in our implementation. Then, we obtain refined motion axes and motion parts using an iteration algorithm. Inspired by Non-Maximum Suppression, we first set the refined motion axis set, $\mathcal{S}$, as empty. We assign the trajectories to candidate axes by computing $dis_{match}$, which lower $dis_{match}$ means that the matching degree is high. Then, we choose the candidate axis that has the most votes. If $\mathcal{S}$ is empty, we add this candidate axis to $\mathcal{S}$. If $\mathcal{S}$ is not empty, we compute whether trajectories belonging to this candidate axis can also belong to an axis in $\mathcal{S}$. If not, we add this candidate axis to $\mathcal{S}$. We iterate these steps until all trajectories are assigned to a motion axis. We optimize coarse segments by examining whether the label of a trajectory is the same as most of the labels of its $k$-nearest neighbors.

\begin{figure*}[t]
\begin{center}
\includegraphics[width=1.0\linewidth]{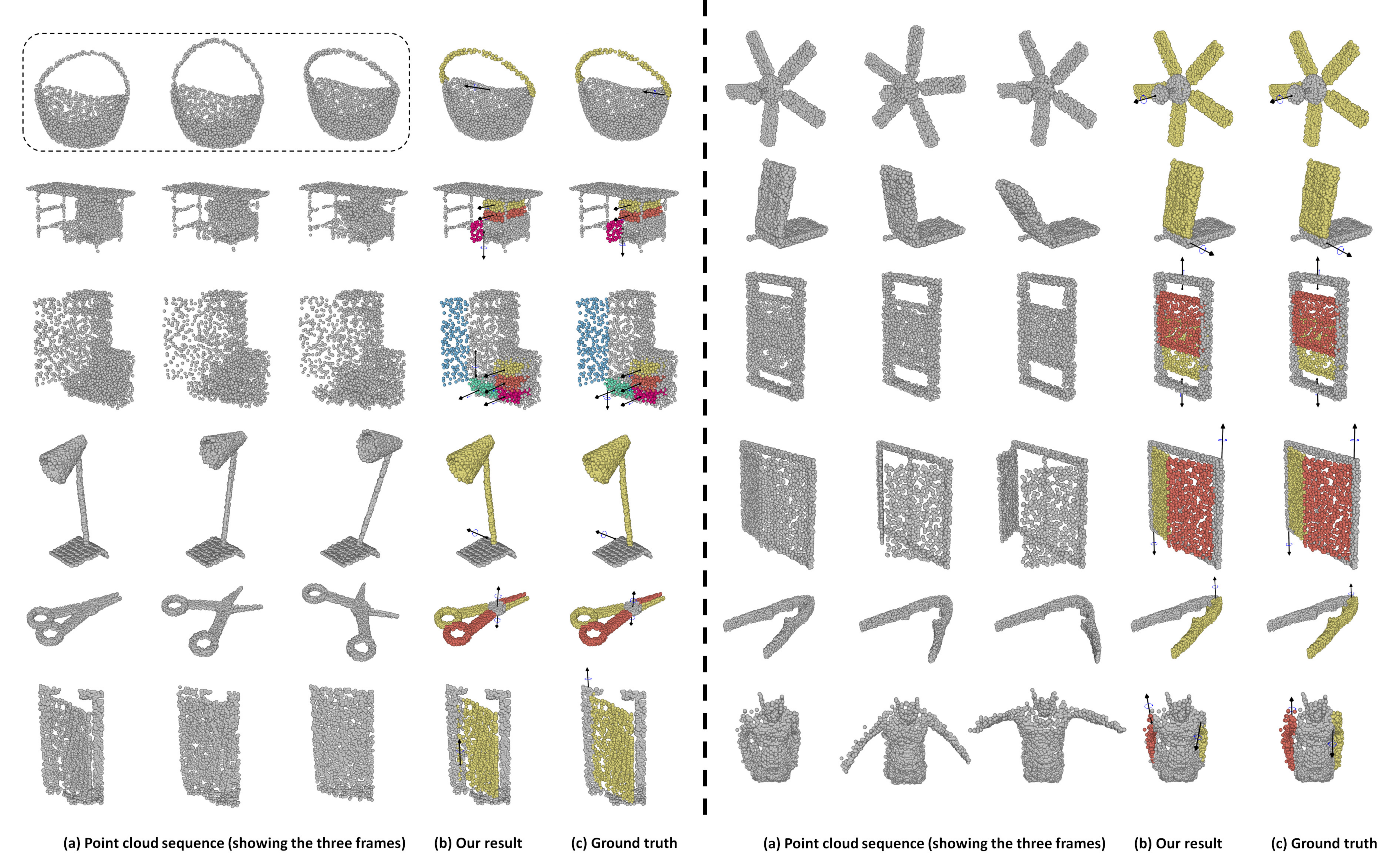}
\end{center}
\caption{Results of our approach compared with the ground truth on both synthetic and real datasets. Rows 1--5 show the results on the synthetic dataset, and row 6 shows the results on the real dataset. We show three frames, first, middle, and lats, of the point cloud sequence. All results are shown in the first frame.}
\label{fig:result}
\end{figure*}

\section{Experiments}
In this section, we first introduce the benchmark datasets used in the experiments (Sec.~\ref{sec:section4_1}). We then compare our approach with state-of-the-art methods on the tasks of motion part segmentation, motion attribute estimation, and motion flow prediction (Sec.~\ref{sec:compare_experiments}).
Finally, we conduct several ablation experiments including the influence of different network designs, the effect of different loss function designs, hyperparameter settings, transferability, etc. (Sec.~\ref{sec:section_4_4}).

\subsection{Dataset Settings}
\label{sec:section4_1}
We evaluate our methods on three different datasets, including two synthetic sets (\textbf{f-data}, \textbf{s-data}) and one real set (\textbf{r-data}). A fake dataset (\textbf{f-data}) is generated randomly and automatically. \textbf{f-data} enables the neural network to learn the common motion patterns of random trajectories. We leverage two annotated datasets including the Motion dataset~\cite{Wang_CVPR2019} and the PartNet dataset~\cite{Mo_2019_CVPR} to construct the second synthetic dataset (\textbf{s-data}). Xiang et al.~\cite{Xiang_2020_SAPIEN} enriched the PartNet dataset with motion attributes. We choose 31 categories including cabinet, lamp, and window, to generate motion sequences. For the real data (\textbf{r-data}), we select certain reasonable motion sequences in the RBO dataset~\cite{RBO_dataset}. Moreover, Yi et al.~\cite{Yi_SIGa2018} provided a small real scan dataset. Additionally, we scan some real data sequences in-house.

\subsection{Comparison with State-of-the-Art Methods}
\label{sec:compare_experiments}
We test our method against four alternatives, including both non-learning and learning approaches. Specifically, we compare our method with the traditional method of Yuan et al.~\cite{Yuan_CGF2016}, and three network-based approaches including Yi et al.~\cite{Yi_SIGa2018}, Wang et al.~\cite{Wang_CVPR2019}, and Yan et al.~\cite{Yan_SIGa19}. 
\subsubsection{Experiment Settings}
To the best of our knowledge, we are the first to propose an approach using deep learning to extract part mobility from a point cloud sequence. To ensure a fair comparison, we implement a space-time co-segmentation baseline following Yuan et al.~\cite{Yuan_CGF2016}. Moreover, we employ the trajectory generation algorithm~\cite{Papazov_and_Burschka_CGF2011} because of their method settings. Yi et al.~\cite{Yi_SIGa2018} proposed a neural network architecture with three modules that propose correspondences, estimate 3D deformation flows, and perform segmentation. Their method input a pair of point clouds representing two different articulation states to segment motion parts and estimate 3D flows. To better provide a contrasting experiment between their approach and ours, we set $t = 2$ in our network settings. We also take a pair of point clouds as input to train and test our network. In terms of the approach of Wang et al.~\cite{Wang_CVPR2019}, we train and evaluate their network and ours using the same training and testing data. However, it must be mentioned that their network requires a single point cloud as input. Considering this difference, the data for their network are randomly sampled frames from point cloud sequences. For the comparison between Yan et al.~\cite{Yan_SIGa19} and our method, the data for their approach are the first frame of the point cloud sequence because their approach segments motion parts by predicting a temporal sequence from a single point cloud.

\begin{table}[t]
\begin{center}

\begin{tabular}{l|c|c|c|c}
\hline
Method  & $IoU$ & MD & OE & TA \\
\hline\hline
Yuan et al.~\cite{Yuan_CGF2016}  & 0.71 & - & - & - \\
\hline
Wang et al.~\cite{Wang_CVPR2019} & 0.67 & 0.051 & 0.055 & 0.92 \\
\hline
Ours & \textbf{0.87} & \textbf{0.032} & \textbf{0.027} & \textbf{0.99} \\
\hline
\end{tabular}
\caption{Comparison between our method and those of Yuan et al.~\cite{Yuan_CGF2016} and Wang et al.~\cite{Wang_CVPR2019} on the tasks of motion part segmentation and motion attribute estimation in terms of $IoU$, MD, OE, and TA.}
\label{tab:compare_with_other_methods}
\end{center}
\end{table}

\subsubsection{Evaluation Metrics}
To evaluate motion part segmentation, we utilize Intersection over Union (\textbf{$IoU$}) defined in~\cite{yi2016scalable}. Moreover, we employ Rand Index (RI) defined in~\cite{2009A} to compare with Yi et al.~\cite{Yi_SIGa2018} and the mean Average Precision (mAP) defined in~\cite{Yan_SIGa19} to compare with~\cite{Yan_SIGa19}. To evaluate motion attribute estimation, we measure the Minimum Distance (MD), Orientation Error (OE), and Type Accuracy (TA), as introduced in~\cite{Wang_CVPR2019}, which are used for measuring the distance, angle, and motion type accuracy between the predicted motion axis line and the ground truth. To evaluate motion flow prediction, we employ 3D end-point-error defined in~\cite{yan2016scene}, which is the average $L_2$ distance between the predicted flow and the ground truth flow.

\begin{table}[t]
\begin{center}
\begin{tabular}{l|c|c|c}
\hline
Method & $IoU$ & RI & EPE\\
\hline\hline
Yi et al.~\cite{Yi_SIGa2018} & 0.71 & 0.80 & 0.029\\
\hline
Ours & \textbf{0.83} & \textbf{0.87} & \textbf{0.024}\\
\hline
\end{tabular}
\end{center}
\caption{Comparison between our method and Yi et al.~\cite{Yi_SIGa2018} on the tasks of motion part segmentation and motion flow prediction in terms of $IoU$, RI, and EPE.}
\label{tab:Compare_with_Yi}
\end{table}

\begin{figure*}[t]
\begin{center}
\includegraphics[width=1.0\linewidth]{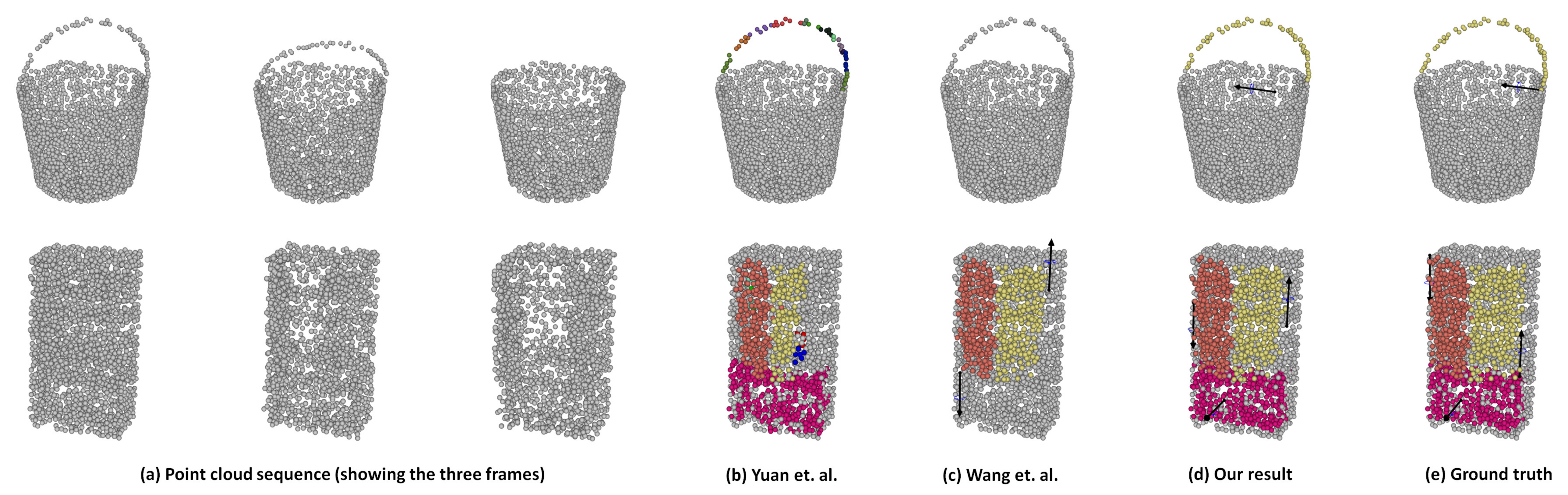}
\end{center}
\caption{Our results are compared with those of Yuan et al.~\cite{Yuan_CGF2016}, and Wang et al.~\cite{Wang_CVPR2019}.}
\label{fig:result}
\end{figure*}

\subsubsection{Results and Analyses}
Yuan et al.~\cite{Yuan_CGF2016} adopted a clustering method to obtain local segments and propagated these segments to the neighboring frames. Finally, they merged all frames using a space--time segment grouping technique to obtain the final motion part segmentation. However, their method tends to have degraded performance when dealing with tiny motion parts. Moreover, their approach has difficulty in processing 3D shapes with many diverse large motions. Tab.~\ref{tab:compare_with_other_methods} shows that our approach outperforms theirs in terms of \textbf{$IoU$}. They did not exploit the fact that motion parts have local characteristics. They generated a motion hypothesis by randomly choosing a number of trajectory triplets. It resulted in the creation of several fragments that were difficult to group together, especially in tiny parts. Additionally, the parameters of their approaches are often different depending on the case.

Wang et al.~\cite{Wang_CVPR2019} parsed 3D shapes into part mobility by observing a single static point cloud based on a large, well-labeled dataset. Their method can only predict motion parameters of 3D shapes that have been seen in the training data because of the supervised learning based on a large, well-annotated dataset. It seems very easy to get confused in extracting motion parts from a single snapshot. For example, it would be challenging to discover whether a door opens on the right or left when it is closed. More importantly, their network is likely to consider that it is not a motion part when a door is closed. They used a similar matrix to segment motion parts. However, this type of method is also not conducive to dealing with tiny parts. In contrast to their method, which directly predicts an axis as a regression problem, we minimize the cosine distance for $\omega$ and solve linear algebraic equations to compute $\mu$, which makes our result more robust and stable than theirs (See Tab.~\ref{tab:compare_with_other_methods}).

\begin{table}[t]
\begin{center}
\begin{tabular}{l|c|c}
\hline
Method & $IoU$ & mAP \\
\hline\hline
Yan et al.~\cite{Yan_SIGa19} &  0.86 & 0.76 \\
\hline
Ours & \textbf{0.87} & \textbf{0.77} \\
\hline
\end{tabular}
\caption{Comparison between our method and Yan et al.~\cite{Yan_SIGa19} on the task of motion part segmentation in terms of $IoU$ and mAP.}
\label{tab:Compare_with_Yan}
\end{center}
\end{table}

Yi et al.~\cite{Yi_SIGa2018} aimed to discover motion parts of objects and estimated 3D flows by analyzing the underlying articulation states and geometry of shapes. Their network architecture alternates between correspondence, deformation flow, and segmentation prediction iteratively in an ICP-like fashion. Theirs is a supervised learning method that estimates 3D point-wise flows and then segments the motion parts. In contrast to their method, we compute the motion axes and motion ranges and then segment motion parts using these motion parameters. The results are reported in Tab.~\ref{tab:Compare_with_Yi}, and they demonstrate that our method achieves higher $IoU$ and RI, and lower EPE than theirs.

Yan et al.~\cite{Yan_SIGa19} predicted a temporal sequence of pointwise displacements from the input shape using LSTM. Then, the RPM-Net used these displacements to learn all the movable points. Next, they obtained a pointwise distance matrix by computing the Euclidean distance between these learned point features. Finally, they selected a clustering method to separate the points in the motion parts according to a distance matrix. They tended to study the part mobility from point cloud sequence using LSTM, but they still used a single point cloud. Tab.~\ref{tab:Compare_with_Yan} shows that the results obtained from these models are close to each other. Although their method deal with motion segmentation well, it has the same defect as that of Wang et al.~\cite{Wang_CVPR2019}, which is due to supervised learning. It is not easy to imagine a motion sequence from a single point cloud.

\subsection{Ablation Experiments}
\label{sec:section_4_4}
In this section, we first conduct several ablation studies to verify the influence of different backbone network designs and different loss function designs. Then, we evaluate our method on different datasets. Finally, we explore the hyperparameter settings in our method and compare them with their model size and timing.

\setlength{\tabcolsep}{4pt}
\begin{table}[t]
\begin{center}
\begin{tabular}{l|c|c|c|c|c|c|c}
\hline
Method  & $IoU$ & MD & OE & $\Theta_{e}$ & $\Phi_{e}$ & $r_{e}$ & TA\\
\hline\hline
baseline1  & 0.86 & 0.029 & 0.054 & 0.054 & 0.042 & 0.025 & 0.97\\
\hline
baseline2  & 0.84 & 0.034 & 0.062 & 0.064 & 0.052 & 0.029 & 0.96\\
\hline
Ours &  \textbf{0.88} & \textbf{0.027} & \textbf{0.028} & \textbf{0.046} & \textbf{0.028} & \textbf{0.023} & \textbf{0.98}\\
\hline
\end{tabular}
\end{center}
\caption{Compare our network with two other network designs on the tasks of motion part segmentation and motion attribute estimation in terms of $IoU$, MD, OE, $\Theta_{e}$, $\Phi_{e}$, $r_{e}$, and TA.}
\label{tab: Compare_with_other_two_network_design}
\end{table}

\subsubsection{Analysis of the Effects of Different Network Designs}

To verify the effectiveness of PointRNN on the tasks of motion part segmentation and motion attribute estimation, we design two baselines to replace the network backbone. To build baseline1, we adopt a slightly modified version of Liu et al.~\cite{Liu_2019_ICCV}. We merge the point cloud sequence into a single point cloud and feed it to a PointNet++ to extract features. Then, we minimize the loss function $\mathcal{L}_{total}$ to estimate motion axis and motion range. Simultaneously, we design baseline2 by referring to Yan et al.~\cite{Yan_SIGa19}. In baseline2, we extract $t$ global features per frame using a shared PointNet++. Then, we adopt an LSTM Network to encode the time information. Similarly, we utilize $\mathcal{L}_{total}$ as the loss function. Different from previous work, our method also estimates the motion ranges including translation distance and rotation angle. Therefore, we three additional novel metrics, $\Theta_{e}$, $\Phi_{e}$, and $r_{e}$. $\Theta_{e}$ is the rotation error between predicted rotation angle and ground truth rotation angle. Similarly, $\Phi_{e}$ is the translation error between predicted translation distance and ground truth translation distance. $r_{e}$ is the error of rebuilding the last frame from the first frame, which is used for a comprehensive evaluation. The results in~\ref{tab: Compare_with_other_two_network_design} demonstrate that our network can achieve better performance than the other two baselines because our network can take advantage of detailed spatiotemporal information from motion sequence.

\begin{table}[t]
\begin{center}
\begin{tabular}{l|c|c}
\hline
\diagbox[width=13em]{Method}{Dataset} & \textbf{f-data} & \textbf{s-data}\\
\hline\hline
Myronenko and Song~\cite{Myronenko_and_Song_arXiv2009} &  \textbf{0.018} & 0.027 \\
\hline
Ours & 0.021 & \textbf{0.023} \\
\hline
\end{tabular}
\end{center}
\caption{Comparison between our method and that of Myronenko and Song~\cite{Myronenko_and_Song_arXiv2009} in terms of $r_e$.}
\label{tab:Compare_with_Myronenko_and_Song}
\end{table}

\subsubsection{Effectiveness Verification of Self-Supervised Method} Our method is self-supervised because our data are not manually annotated, either in training or testing. Moreover, supervised information is automatically generated from the characteristic distribution of the data. To this end, we define a loss function, $\mathcal{L}_{axis}$. We solve an absolute orientation problem to generate a motion matrix ($\hat{\bm{M}}$) using the method of Myronenko and Song~\cite{Myronenko_and_Song_arXiv2009}. Then, we parse the supervised information ($\hat{\mu}$, $\hat{\omega}$, $\hat{\Phi}$, $\hat{\Theta}$) from $\hat{\bm{M}}$. 

In this section, we conduct an ablation study to compare with the performance of~\cite{Myronenko_and_Song_arXiv2009}, since not improving on~\cite{Myronenko_and_Song_arXiv2009} would have meant the neural network architecture would not be necessary. The results in Tab.~\ref{tab:Compare_with_Myronenko_and_Song} show that our method is better than~\cite{Myronenko_and_Song_arXiv2009} on \textbf{s-data}, but worse on \textbf{f-data}. It demonstrates a hypothesis that our method can handle more complex data. \textbf{s-data} usually contains two or more motion parts and a static part. However, \textbf{f-data} only has one motion part. Obviously, \textbf{s-data} is more complex than \textbf{f-data}. The output of~\cite{Myronenko_and_Song_arXiv2009} is an analytical solution. If the grouped trajectories cover different motion parts, it will output the wrong answer. However, our loss function contains three components instead of only utilizing $\mathcal{L}_{axis}$, and the deep learning method learns common motion features and outputs an approximation solution. It might slightly alleviate the problem of the wrong solution on \textbf{s-data}.

\setlength{\tabcolsep}{4pt}
\begin{table}[t]
\begin{center}
\begin{tabular}{l|c|c|c|c|c|c}
\hline
Loss Function& MD & OE & $\Theta_{e}$ & $\Phi_{e}$ & $r_{e}$ & TA \\
\hline\hline
\makecell[l]{$\mathcal{L}_{axis}$}-\textbf{f} & 0.035 & 0.017 & 0.058 & 0.026 & 0.061 & 0.96 \\
\hline
\makecell[l]{$\mathcal{L}_{rebuild} +\mathcal{L}_{pose}$ }-\textbf{f} & 0.031 & 0.017 & 0.051 & 0.036 & 0.086 & 0.90 \\
\hline
\makecell[l]{$\mathcal{L}_{pose} +\mathcal{L}_{axis}$}-\textbf{f} & 0.032 & 0.016 & 0.053 & \textbf{0.022} & 0.045 & 0.94\\
\hline
\makecell[l]{$\mathcal{L}_{total}$}-\textbf{f} & \textbf{0.027} & \textbf{0.014} & \textbf{0.047} & 0.023 & \textbf{0.020} & \textbf{0.97} \\
\hline\hline

\makecell[l]{$\mathcal{L}_{axis}$}-\textbf{s} & 0.043 & 0.030 & 0.056 & 0.030 & 0.045 & 0.97 \\
\hline
\makecell[l]{$\mathcal{L}_{rebuild} +\mathcal{L}_{pose}$ }-\textbf{s} & 0.048 & 0.033 & 0.057 & 0.034 & 0.061 & 0.93\\
\hline
\makecell[l]{$\mathcal{L}_{pose} +\mathcal{L}_{axis}$}-\textbf{s} & 0.037 & 0.029 & 0.054 & 0.030 & 0.034 & \textbf{0.99}\\
\hline
\makecell[l]{$\mathcal{L}_{total}$}-\textbf{s} & \textbf{0.032} & \textbf{0.027} & \textbf{0.046} & \textbf{0.027} & \textbf{0.023} & \textbf{0.99} \\
\hline

\end{tabular}
\end{center}
\caption{Effectiveness of different loss function designs on the task of motion attribute estimation. The first four rows show the results on \textbf{f-data}, and the last four rows show the results on \textbf{s-data}.}
\label{tab:loss_design}
\end{table}

\subsubsection{Effect of Different Loss Function Designs} To train our network using self-supervised learning, we define a novel self-supervised loss function including three components. To analyze the influence of different loss function designs, we experiment with an ablated version of our network with four combinations of all loss functions on two datasets including \textbf{f-data} and \textbf{s-data}. We utilize the loss function to train the network on the task of motion hypotheses generation. Therefore, different loss designs mainly influence the network performance with respect to axis generation. Thus, we train and test our network on the task of motion attribute estimation to focus on examining the quality of the predicted motion axes. Tab.~\ref{tab:loss_design} lists the results of including $\mathcal{L}_{axis}$, $\mathcal{L}_{rebuild} + \mathcal{L}_{pose}$, $\mathcal{L}_{pose} + \mathcal{L}_{axis}$, and $\mathcal{L}_{total}$. The results demonstrate that our method achieves optimal performance for motion axis estimation, thereby verifying its optimal effect when using $\mathcal{L}_{rebuild}$, $\mathcal{L}_{pose}$, and $\mathcal{L}_{axis}$ together. Moreover, the shown results are applicable to both \textbf{f-data} and \textbf{s-data}. The reason for this is that it is possible to generate the same $\bm{M}$ using two different parameter sets of $\mu$, $\omega$, $\Phi$, and $\Theta$. There is an interactive constraint between different components of the loss function.

\setlength{\tabcolsep}{4pt}
\begin{table}[t]
\begin{center}
\begin{tabular}{l|c|c|c|c|c|c|c}
\hline
Dataset & $IoU$ & MD & OE & $\Theta_{e}$ & $\Phi_{e}$ & $r_{e}$ & TA \\
\hline\hline
\textbf{f-data} & - & \textbf{0.027} & \textbf{0.014} & 0.047 & \textbf{0.023} & \textbf{0.020} & 0.97\\
\hline
\textbf{s-data} & \textbf{0.87} & 0.032 & 0.027 & \textbf{0.046} & 0.027 & 0.023 & \textbf{0.99}\\
\hline
\textbf{r-data} & 0.79 & 0.058 & 0.036 & 0.079 & 0.037 & 0.058 & \textbf{0.99}\\
\hline
\end{tabular}
\end{center}
\caption{Performance of our method on three datasets on the tasks of motion part segmentation and motion attribute estimation in terms of $IoU$, MD, OE, $\Theta_{e}$, $\Phi_{e}$, $r_{e}$, and TA.}
\label{tab:three_dataset_result}
\end{table} 

\subsubsection{Performance on Three Datasets} We have three datasets used in experiments, and these datasets have their own unique features. The difficulty of \textbf{f-data} stems from its randomness. Each element of \textbf{f-data} has totally different motion parameters, including motion axes and motion ranges. \textbf{s-data} has various categories. Most 3D shapes in \textbf{s-data} have two or more motion parts. \textbf{r-data} generally has a single partial view of 3D shapes with more noise. Therefore, we need to verify the performance on these datasets, respectively. We train and test our network using three different datasets. The results are reported in Tab.~\ref{tab:three_dataset_result}, and results demonstrate that not only can our approach learn motion parameters from synthetic data, but it can also be applied to real noisy data.

\begin{figure}[t]
\centering
\includegraphics[width=\linewidth]{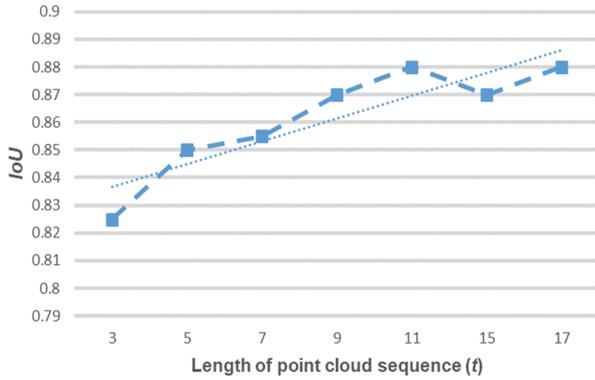}
\caption{Impact of the length of a point cloud sequence, where a higher $IoU$ value is better. The x-axis corresponds to the length of the point cloud sequence ($t$).}
\label{fig:t_k_analysis}
\end{figure}

\begin{table}[t]
\begin{center}
\begin{tabular}{l|c|c|c|c|c|c}
\hline
category & $IoU$ & MD & OE & $\Theta_{e}$ & $r_{e}$ & TA\\
\hline\hline
\makecell[l]{Fan} -\textbf{f} & 0.89 & 0.014 & 0.014 & 0.032 & 0.028 & 1\\
\hline
\makecell[l]{Laptop} -\textbf{f} & 0.95 & 0.018 & 0.013 & 0.035 & 0.021 & 1\\
\hline
\makecell[l]{Scissor} -\textbf{f} & 0.79 & 0.022 & 0.017 & 0.087 & 0.031 & 1\\
\hline\hline
\makecell[l]{Fan} -\textbf{s} & 0.93 & 0.012 & 0.014 & 0.028 & 0.024 & 1\\
\hline
\makecell[l]{Laptop} -\textbf{s} & 0.97 & 0.017 & 0.014 & 0.031 & 0.017 & 1\\
\hline
\makecell[l]{Scissor} -\textbf{s} & 0.84 & 0.019 & 0.015 & 0.069 & 0.028 & 1\\
\hline
\end{tabular}
\end{center}
\caption{Experiment on the motion feature learning. The first three rows show the results, which train our network on \textbf{f-data}, and the last three rows show the results, which train our network on \textbf{s-data}. Our method can handle unseen categories, because motion patterns and type are learned by the network.}
\label{tab:generalization}
\end{table}

\subsubsection{Analysis on Motion Feature Learning} Supervised learning methods tend to require that object categories in test data must have been seen in train data. However, our approach is a self-supervised deep learning algorithm, and our network can estimate the motion axes by learning the feature representation of trajectories other than only by learning geometrical characteristics. Therefore, we aim to verify that our network has the ability to learn motion features. We designed an experiment in which we train our network on \textbf{f-data} and \textbf{s-data}, and then test it on \textbf{s-data}. We present three categories with different levels of complexity results in Tab. ~\ref{tab:generalization}. Two sets of the result that train our network on the different datasets are similar. The results demonstrate that the network can produce correct motion axes, even though it did not see any real objects. It implies that our network can learn motion patterns from trajectories.

\subsubsection{Analysis on the Length of Point Cloud Sequence}
Considering that our motivation is to learn part mobility from a point cloud sequence, the number of frames ($t$) is an important hyperparameter in our approach. Thus, we must determine how many frames are sampled from one motion sequence. To achieve this, we conduct an ablation experiment on $t$ to discuss the impact of the length of the point cloud sequence. We reduce the number of frames from 17 to 3 for sufficient verification. The results are depicted in Fig.~\ref{fig:t_k_analysis}. As the number of frames increases, the performance gain is consistent but slower and slower. Hence, we adopt 11 frames as input to balance performance and efficiency.

\subsubsection{Effect of Trajectory Aggregation} 
\begin{figure}[t]
\centering
\includegraphics[width=\linewidth]{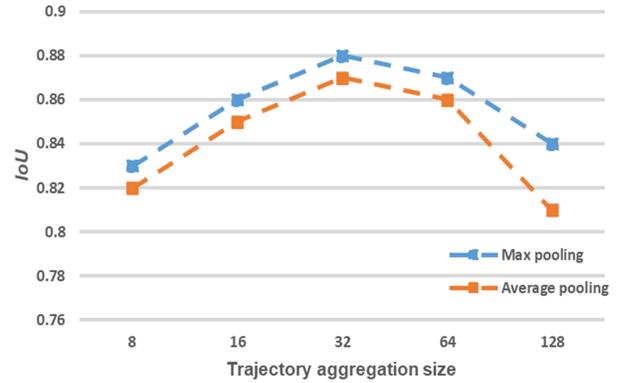}
\caption{Trajectory aggregation analysis, where a higher $IoU$ value is better. The x-axis corresponds to different hyperparameters of trajectory aggregation size ($k$).}
\label{fig:aggregation_analysis}
\end{figure}

The aggregation of trajectories is of great importance in PointRNN, as it describes a motion part consisting of a group of similar motion trajectories. Thus, it is important to analyze how different aggregation designs influence performance. We report five different sizes with two types of aggregations: max and average, as shown in Fig.~\ref{fig:aggregation_analysis}. As the aggregation radius ($k$) increases, PointRNN improves until it peaks at approximately 32. The results demonstrate that it is challenging to learn the features when clustering too few trajectories in a local region. In contrast, a larger region usually contains more than one motion part and results in decreased performance. Moreover, the max-pooling operation achieves better results than average by aggregating the features in local regions. 

\begin{figure}[t]
\centering
\includegraphics[width=\linewidth]{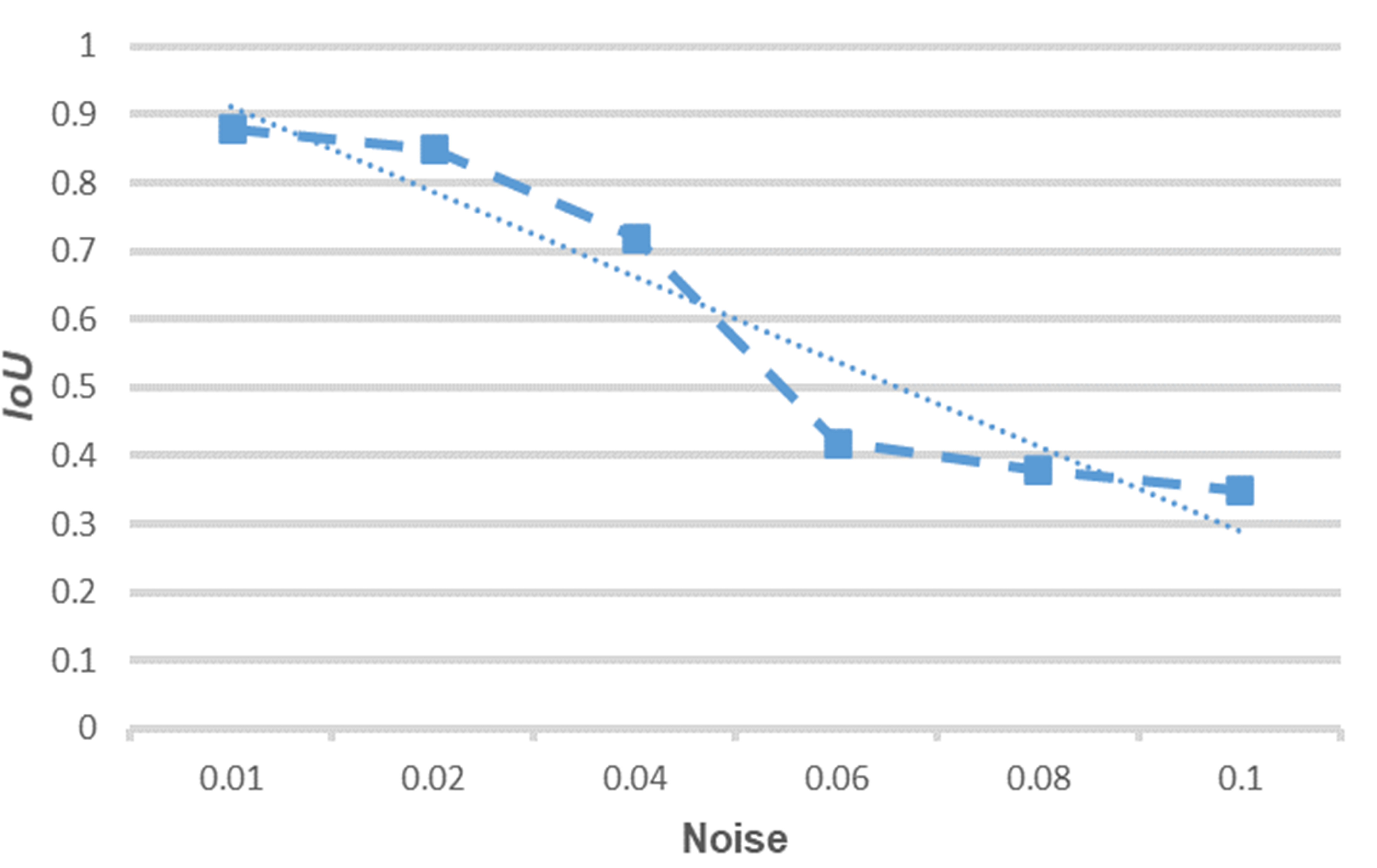}
\caption{Effect of different qualities of trajectory. The result illustrates that the performance decreases with increasing of noise.}
\label{fig:quality}
\end{figure}

\subsubsection{Analysis on the Quality of Trajectory} We also discuss the impact of different qualities of trajectory. Given a point cloud sequence, we first generate a set of trajectories. Then, we add different levels of random noise to these trajectories. Fig.~\ref{fig:quality} shows the relationship between \textbf{$IoU$} and noise. Our method maintained more than 0.7 (\textbf{$IoU$}) after adding 0.04 random noise. As the noise degree increases, the performance declines sharply. The results demonstrate that our method can tolerate noise to a certain extent, but it will fail when under the addition of excessive noise.

\begin{table}[t]
\begin{center}
\begin{tabular}{l|c|c}
\hline
Method & Model size & Timing\\
\hline\hline
Yuan et al.~\cite{Yuan_CGF2016} & - & -/180s\\
\hline
Wang et al.~\cite{Wang_CVPR2019} & 86.7MB & 38h/7.32s\\
\hline
Yi et al.~\cite{Yi_SIGa2018} & 78MB & 35.5h/4.88s\\
\hline
Yan et al.~\cite{Yan_SIGa19} & 153.3 MB & 13.2h/ 0.55s\\
\hline
Ours & \textbf{25.2MB} & \textbf{11.6h}/\textbf{0.43s}\\
\hline
\end{tabular}
\end{center}
\caption{Comparison of Model size and processing time. Our model is smaller and faster.}
\label{tab:Model_size_and_processing_time}
\end{table}

\subsubsection{Model Size and Speed}
Our proposed model is highly efficient (See Tab.~\ref{tab:Model_size_and_processing_time}), as it leverages sparsity in point clouds by using clusters and does not require many stages like those of Wang et al.~\cite{Wang_CVPR2019} and Yi et al. ~\cite{Yi_SIGa2018}. Compared to previous methods, our model is more than $3\times$ smaller in size and more than $10\times$ times faster than those of Wang et al.~\cite{Wang_CVPR2019}, and Yi et al. ~\cite{Yi_SIGa2018}. While the method in Yuan et al.~\cite{Yuan_CGF2016} is a non-learning method, it costs approximately 3 min to segment a 3D shape without giving motion parameters. Furthermore, our model is more than $6\times$ smaller in size than that of Yan et al.~\cite{Yan_SIGa19}.  Moreover, we employ the model size to compare the size of the training parameters of different models to evaluate the memory consumption of the algorithm.

\section{Limitations and Future Work}
There are several avenues for future research. First, our method is based on trajectories such that if the quality of trajectories is extremely poor with many fragments, the performance decreases. Second, if two parts have exactly the same motion, e.g., two drawers in a cabinet have the same motion, our method considers that they have the same motion and should be one motion part. Third, for hierarchical mobility extraction, we can segment the 3D object into different motion parts, but the motion axis describes a compound movement. For example, the motion of a bulb holder is a composite movement with that of a lamp post. In the future, it would be interesting to infer parts and motions and discover common articulation patterns from various sensor data. Moreover, our method relies on a pre-registration of point clouds using an ICP-like optimization. There are also many deep learning methods for point cloud registration, such as~\cite{Pais20203DRegNetAD}. It is deserved to design a network encompassing both trajectory generation and motion attribute estimation in an end-to-end fashion in future work. To facilitate future research and reproduce our method more easily, the source code and dataset are available on Github: \url{https://github.com/AGithubRepository/PartMobility}.

\section{Conclusion}
In this paper, we introduce a self-supervised method for parsing 3D shapes into several motion parts with motion parameters from point cloud sequences. We first transform point cloud sequences into trajectories. Then, we propose a novel neural network architecture called PointRNN to extract the feature representations from trajectories and produce motion hypotheses. Finally, we propose an iterative algorithm for segmenting motion parts. In contrast to previous studies, our approach first predicts part motion parameters including motion axes and motion ranges and then segments motion parts guided by these motion hypotheses. We experimentally demonstrated that our approach yields significantly better results compared with both the traditional and network-based methods.

\section*{Acknowledgement}
We thank the anonymous reviewers for their valuable comments. This work was supported in part by National Key Research and Development Program of China (2018YFC0831003 and 2019YFF0302902), National Natural Science Foundation of China (61902014 and U1736217 and 61932003), and Pre-research Project of the Manned Space Flight (060601).

\bibliographystyle{eg-alpha-doi}

\bibliography{egbibsample}

\newcommand{\etalchar}[1]{$^{#1}$}
\begin{thebibliography}{\uppercase{HLVK{\etalchar{*}}17}}

\bibitem[Bel]{RodriguesRotationFormula}
\textsc{Belongie S.}:
\newblock Rodrigues' rotation formula.
\newblock From MathWorld--A Wolfram Web Resource, created by Eric W. Weisstein.
  https://mathworld.wolfram.com/RodriguesRotationFormula.html.

\bibitem[BM92]{BM92}
\textsc{Besl P.~J., McKay N.~D.}:
\newblock A method for registration of 3-d shapes.
\newblock \emph{IEEE Trans. Pattern Anal. Mach. Intell. 14}, 2 (Feb. 1992),
  239–256.

\bibitem[BPDG19]{Behl_CVPR2019}
\textsc{Behl A., Paschalidou D., Donne S., Geiger A.}:
\newblock Pointflownet: Learning representations for rigid motion estimation
  from point clouds.
\newblock In \emph{The IEEE Conference on Computer Vision and Pattern
  Recognition (CVPR)} (June 2019).

\bibitem[CGF09]{2009A}
\textsc{Chen X., Golovinskiy A., Funkhouser T.}:
\newblock A benchmark for 3d mesh segmentation.
\newblock \emph{Acm Transactions on Graphics 28}, 3 (2009), 1--12.

\bibitem[CZ08]{Chang_and_Zwicker_CGF2008}
\textsc{Chang W., Zwicker M.}:
\newblock Automatic registration for articulated shapes.
\newblock \emph{Computer Graphics Forum 27}, 5 (2008), 1459--1468.

\bibitem[FRA11]{Fayad_ICCV2011}
\textsc{{Fayad} J., {Russell} C., {Agapito} L.}:
\newblock Automated articulated structure and 3d shape recovery from point
  correspondences.
\newblock In \emph{2011 International Conference on Computer Vision} (Nov
  2011), pp.~431--438.

\bibitem[HLRB13]{Hermans_2013}
\textsc{{Hermans} T., {Li} F., {Rehg} J.~M., {Bobick} A.~F.}:
\newblock Learning contact locations for pushing and orienting unknown objects.
\newblock In \emph{2013 13th IEEE-RAS International Conference on Humanoid
  Robots (Humanoids)} (Oct 2013), pp.~435--442.

\bibitem[HLVK{\etalchar{*}}17]{Hu_SIGa2017}
\textsc{Hu R., Li W., Van~Kaick O., Shamir A., Zhang H., Huang H.}:
\newblock Learning to predict part mobility from a single static snapshot.
\newblock \emph{ACM Trans. Graph. 36}, 6 (Nov. 2017), 227:1--227:13.

\bibitem[HS97]{LSTM}
\textsc{Hochreiter S., Schmidhuber J.}:
\newblock Long short-term memory.
\newblock \emph{Neural Computation 9}, 8 (1997), 1735--1780.

\bibitem[HSvK18]{Hu_CGF2018}
\textsc{Hu R., Savva M., van Kaick O.}:
\newblock Functionality representations and applications for shape analysis.
\newblock \emph{Computer Graphics Forum 37}, 2 (2018), 603--624.

\bibitem[JSGC15]{Jaimez_ICRA2015}
\textsc{{Jaimez} M., {Souiai} M., {Gonzalez-Jimenez} J., {Cremers} D.}:
\newblock A primal-dual framework for real-time dense rgb-d scene flow.
\newblock In \emph{2015 IEEE International Conference on Robotics and
  Automation (ICRA)} (May 2015), pp.~98--104.

\bibitem[KL17]{Klokov_ICCV2017}
\textsc{Klokov R., Lempitsky V.}:
\newblock Escape from cells: Deep kd-networks for the recognition of 3d point
  cloud models.
\newblock In \emph{The IEEE International Conference on Computer Vision (ICCV)}
  (Oct 2017).

\bibitem[KLAK16]{Kim_IROS2016}
\textsc{{Kim} Y., {Lim} H., {Ahn} S.~C., {Kim} A.}:
\newblock Simultaneous segmentation, estimation and analysis of articulated
  motion from dense point cloud sequence.
\newblock In \emph{2016 IEEE/RSJ International Conference on Intelligent Robots
  and Systems (IROS)} (Oct 2016), pp.~1085--1092.

\bibitem[LBS{\etalchar{*}}18]{PointCNN}
\textsc{Li Y., Bu R., Sun M., Wu W., Di X., Chen B.}:
\newblock Pointcnn: Convolution on x-transformed points.
\newblock In \emph{Advances in Neural Information Processing Systems 31}.
  Curran Associates, Inc., 2018, pp.~820--830.

\bibitem[LQG19]{Liu_CVPR2019}
\textsc{Liu X., Qi C.~R., Guibas L.~J.}:
\newblock Flownet3d: Learning scene flow in 3d point clouds.
\newblock In \emph{The IEEE Conference on Computer Vision and Pattern
  Recognition (CVPR)} (June 2019).

\bibitem[LSZ{\etalchar{*}}19]{Liu_2019_CVM}
\textsc{Liu M., Shi Y., Zheng L., Xu K., Huang H., Manocha D.}:
\newblock Recurrent 3d attentional networks for end-to-end active object
  recognition.
\newblock \emph{Computational Visual Media 5}, 01 (2019), 92--104.

\bibitem[LWL{\etalchar{*}}16]{Li_CGF2016}
\textsc{Li H., Wan G., Li H., Sharf A., Xu K., Chen B.}:
\newblock Mobility fitting using 4d ransac.
\newblock \emph{Computer Graphics Forum 35}, 5 (2016), 79--88.

\bibitem[LYB19]{Liu_2019_ICCV}
\textsc{Liu X., Yan M., Bohg J.}:
\newblock Meteornet: Deep learning on dynamic 3d point cloud sequences.
\newblock In \emph{The IEEE International Conference on Computer Vision (ICCV)}
  (October 2019).

\bibitem[MMEB18]{RBO_dataset}
\textsc{Martln-Martln R., Eppner C., Brock O.}:
\newblock The rbo dataset of articulated objects and interactions, 2018.

\bibitem[MS09]{Myronenko_and_Song_arXiv2009}
\textsc{Myronenko A., Song X.~B.}:
\newblock On the closed-form solution of the rotation matrix arising in
  computer vision problems.
\newblock \emph{CoRR abs/0904.1613} (2009).

\bibitem[MS15]{Maturana_and_Scherer_ICRA2015}
\textsc{{Maturana} D., {Scherer} S.}:
\newblock 3d convolutional neural networks for landing zone detection from
  lidar.
\newblock In \emph{2015 IEEE International Conference on Robotics and
  Automation (ICRA)} (May 2015), pp.~3471--3478.

\bibitem[MTFA15]{Myers_ICRA2015}
\textsc{{Myers} A., {Teo} C.~L., {Fermller} C., {Aloimonos} Y.}:
\newblock Affordance detection of tool parts from geometric features.
\newblock In \emph{2015 IEEE International Conference on Robotics and
  Automation (ICRA)} (May 2015), pp.~1374--1381.

\bibitem[MYY{\etalchar{*}}13]{Mitra_2013}
\textsc{Mitra N.~J., Yang Y.-L., Yan D.-M., Li W., Agrawala M.}:
\newblock Illustrating how mechanical assemblies work.
\newblock \emph{Commun. ACM 56}, 1 (Jan. 2013), 106--114.

\bibitem[MZC{\etalchar{*}}19]{Mo_2019_CVPR}
\textsc{Mo K., Zhu S., Chang A.~X., Yi L., Tripathi S., Guibas L.~J., Su H.}:
\newblock {PartNet}: A large-scale benchmark for fine-grained and hierarchical
  part-level {3D} object understanding.
\newblock In \emph{The IEEE Conference on Computer Vision and Pattern
  Recognition (CVPR)} (June 2019).

\bibitem[PB11]{Papazov_and_Burschka_CGF2011}
\textsc{Papazov C., Burschka D.}:
\newblock Deformable 3d shape registration based on local similarity
  transforms.
\newblock \emph{Computer Graphics Forum 30}, 5 (2011), 1493--1502.

\bibitem[PMR{\etalchar{*}}20]{Pais20203DRegNetAD}
\textsc{Pais G., Miraldo P., Ramalingam S., Govindu V., Nascimento J.,
  Chellappa R.}:
\newblock 3dregnet: A deep neural network for 3d point registration.
\newblock \emph{2020 IEEE/CVF Conference on Computer Vision and Pattern
  Recognition (CVPR)} (2020), 7191--7201.

\bibitem[QYSG17]{Pointnet++}
\textsc{Qi C.~R., Yi L., Su H., Guibas L.~J.}:
\newblock Pointnet++: Deep hierarchical feature learning on point sets in a
  metric space.
\newblock In \emph{Advances in Neural Information Processing Systems 30}.
  Curran Associates, Inc., 2017, pp.~5099--5108.

\bibitem[RLA{\etalchar{*}}19]{Roberts_2019_CVM}
\textsc{Roberts R., Lewis J.~P., Anjyo K., Seo J., Seol Y.}:
\newblock Optimal and interactive keyframe selection for motion capture.
\newblock \emph{Computational Visual Media 5}, 02 (2019), 172--191.

\bibitem[SA07]{ARAP}
\textsc{Sorkine O., Alexa M.}:
\newblock As-rigid-as-possible surface modeling, 2007.

\bibitem[SMW06]{Scott_SIGGRAPH2006}
\textsc{Schaefer S., McPhail T., Warren J.}:
\newblock Image deformation using moving least squares.
\newblock \emph{ACM Trans. Graph. 25} (07 2006), 533--540.

\bibitem[SSTN18]{Supasorn_NIPS2018}
\textsc{Suwajanakorn S., Snavely N., Tompson J.~J., Norouzi M.}:
\newblock Discovery of latent 3d keypoints via end-to-end geometric reasoning.
\newblock In \emph{Advances in Neural Information Processing Systems 31}.
  Curran Associates, Inc., 2018, pp.~2059--2070.

\bibitem[VRS14]{Vogel_ECCV2014}
\textsc{Vogel C., Roth S., Schindler K.}:
\newblock View-consistent 3d scene flow estimation over multiple frames.
\newblock In \emph{Computer Vision -- ECCV 2014} (Cham, 2014), Springer
  International Publishing, pp.~263--278.

\bibitem[WZS{\etalchar{*}}19]{Wang_CVPR2019}
\textsc{Wang X., Zhou B., Shi Y., Chen X., Zhao Q., Xu K.}:
\newblock Shape2motion: Joint analysis of motion parts and attributes from 3d
  shapes.
\newblock In \emph{The IEEE Conference on Computer Vision and Pattern
  Recognition (CVPR)} (June 2019).

\bibitem[XQM{\etalchar{*}}20]{Xiang_2020_SAPIEN}
\textsc{Xiang F., Qin Y., Mo K., Xia Y., Zhu H., Liu F., Liu M., Jiang H., Yuan
  Y., Wang H., Yi L., Chang A.~X., Guibas L.~J., Su H.}:
\newblock {SAPIEN}: A simulated part-based interactive environment.
\newblock In \emph{The IEEE Conference on Computer Vision and Pattern
  Recognition (CVPR)} (June 2020).

\bibitem[YHL{\etalchar{*}}18]{Yi_SIGa2018}
\textsc{Yi L., Huang H., Liu D., Kalogerakis E., Su H., Guibas L.}:
\newblock Deep part induction from articulated object pairs.
\newblock \emph{ACM Trans. Graph. 37}, 6 (Dec. 2018), 209:1--209:15.

\bibitem[YHY{\etalchar{*}}19]{Yan_SIGa19}
\textsc{Yan Z., Hu R., Yan X., Chen L., van Kaick O., Zhang H., Huang H.}:
\newblock Rpm-net: Recurrent prediction of motion and parts from point cloud.
\newblock \emph{ACM Transactions on Graphics (Proceedings of SIGGRAPH ASIA
  2019) 38}, 6 (2019), 240:1--240:15.

\bibitem[YKC{\etalchar{*}}16]{yi2016scalable}
\textsc{Yi L., Kim V.~G., Ceylan D., Shen I.-C., Yan M., Su H., Lu C., Huang
  Q., Sheffer A., Guibas L.}:
\newblock A scalable active framework for region annotation in 3d shape
  collections.
\newblock \emph{ACM Transactions on Graphics (ToG) 35}, 6 (2016), 1--12.

\bibitem[YLX{\etalchar{*}}16]{Yuan_CGF2016}
\textsc{Yuan Q., Li G., Xu K., Chen X., Huang H.}:
\newblock Space-time co-segmentation of articulated point cloud sequences.
\newblock \emph{Computer Graphics Forum 35}, 2 (2016), 419--429.

\bibitem[YP06]{Yan_ECCV2006}
\textsc{Yan J., Pollefeys M.}:
\newblock A general framework for motion segmentation: Independent,
  articulated, rigid, non-rigid, degenerate and non-degenerate.
\newblock In \emph{Computer Vision -- ECCV 2006} (Berlin, Heidelberg, 2006),
  Springer Berlin Heidelberg, pp.~94--106.

\bibitem[YX16]{yan2016scene}
\textsc{Yan Z., Xiang X.}:
\newblock Scene flow estimation: A survey.
\newblock \emph{arXiv preprint arXiv:1612.02590} (2016).

\end{thebibliography}


\end{document}